%% file: root.tex
\pdfoutput=1

\documentclass[letterpaper, 10 pt, conference]{ieeeconf}  

\IEEEoverridecommandlockouts                              

\usepackage{hyperref}
\hypersetup{
  pdftitle={RoboMorph: Evolving Robot Morphology using Large Language Models},
  pdfauthor={Kevin Qiu, Wladyslaw Palucki, Krzysztof Ciebiera, Pawel Fijalkowski, Marek Cygan, Lukasz Kucinski},
  colorlinks=true,
  linkcolor=blue,
  citecolor=blue,
  urlcolor=blue,
}
\usepackage{url}
\usepackage{algorithmic}
\usepackage{algorithm}
\usepackage{mdframed}
\usepackage{graphicx}
\usepackage{subcaption}
\let\labelindent\relax
\usepackage{enumitem} 
\usepackage{wrapfig}
\usepackage{booktabs}
\usepackage[table]{xcolor}
\usepackage{cite}

\newcommand{\method}{\texttt{RoboMorph}}
\newcommand{\kq}[1]{} 

\title{\LARGE \bf
\method{}: Evolving Robot Morphology using Large Language Models
}

\author{Kevin Qiu$^{1,2}$, W{\l}adys{\l}aw Pa{\l}ucki$^{1}$, Krzysztof Ciebiera$^{1}$, Pawe{\l} Fija{\l}kowski$^{1}$,\\
Marek Cygan$^{1,3}$, {\L}ukasz Kuci\'{n}ski$^{1,2,4}$\\[0.5em]
{\normalsize $^{1}$University of Warsaw \quad $^{2}$IDEAS NCBR \quad $^{3}$Nomagic \quad $^{4}$Polish Academy of Sciences}\\
{\normalsize\texttt{kevinxqiu@gmail.com}}
}

\begin{document}
\bstctlcite{IEEEexample:BSTcontrol}

\maketitle


\begin{abstract}
We introduce \method{}, an automated approach for generating and optimizing modular robot designs using large language models (LLMs) and evolutionary algorithms. Each robot design is represented by a structured grammar, and we use LLMs to efficiently explore this design space. Traditionally, such exploration is time-consuming and computationally intensive. Using a best-shot prompting strategy combined with reinforcement learning (RL)-based control evaluation, \method{} iteratively refines robot designs within an evolutionary feedback loop. Across four terrain types, \method{} discovers diverse, terrain-specialized morphologies, including wheeled quadrupeds and hexapods, that match or outperform designs produced by Robogrammar's graph-search method. These results demonstrate that LLMs, when coupled with evolutionary selection, can serve as effective generative operators for automated robot design. Our project page and code are available at \url{https://robomorph.github.io}.
\end{abstract}

\input{sections/intro}
\input{sections/method}
\input{sections/results}
\input{sections/related_work}
\input{sections/future_work}

\input{sections/conclusion}

\input{sections/acknowledgements}

\bibliographystyle{IEEEtran}
\input{root.bbl}


\end{document}

%% file: sections/intro.tex
\section{Introduction}
\label{sec:intro}

Designing robot morphologies remains a fundamental challenge in modern robotics \cite{zeng2023large}. Traditional engineering approaches are time-consuming, heavily dependent on human intuition, and ill-suited to exploring the vast space of possible designs. Automating this process could significantly reduce development time and uncover solutions that surpass those conceived by human designers.

Large language models (LLMs) have recently transformed multiple areas of robotics, from generating robot policies as code \cite{liang2023code} and enhancing semantic reasoning \cite{brohan2023rt1} to designing reward functions and domain randomization \cite{ma2024eureka, ma2024dreureka}. Here, we apply these capabilities to the robot design problem itself.

Previous attempts to automate the design process have been computationally expensive, requiring extensive search over large design spaces. One class of such approaches is inspired by evolutionary algorithms (EAs)~\cite{back1993overview}. However, EA-based methods require hundreds of generations and struggle to scale with increasing design complexity~\cite{thierens1999scalability}. Other methods, such as Robogrammar~\cite{zhao2020robogrammar}, introduce structured grammars and heuristic search, while recent latent optimization methods~\cite{hu2023glso, xu2021multi} focus on guiding exploration with learned representations. Despite their successes, these approaches remain limited by fixed heuristics, narrow exploration, or computational bottlenecks.

We introduce \method{}, a framework that integrates LLM-based generation, evolutionary algorithms, robot grammars, and RL-based control evaluation. Unlike prior work, \method{} uses the compositional priors of LLMs as a mutation operator, enabling the discovery of morphologies that classical search heuristics overlook. By exploiting the reasoning capabilities of LLMs within the grammar, our method accelerates exploration of the design space and uncovers unconventional solutions beyond what human-engineered heuristics or latent models produce.

\begin{figure}
    \centering
    \includegraphics[width=0.7\columnwidth]{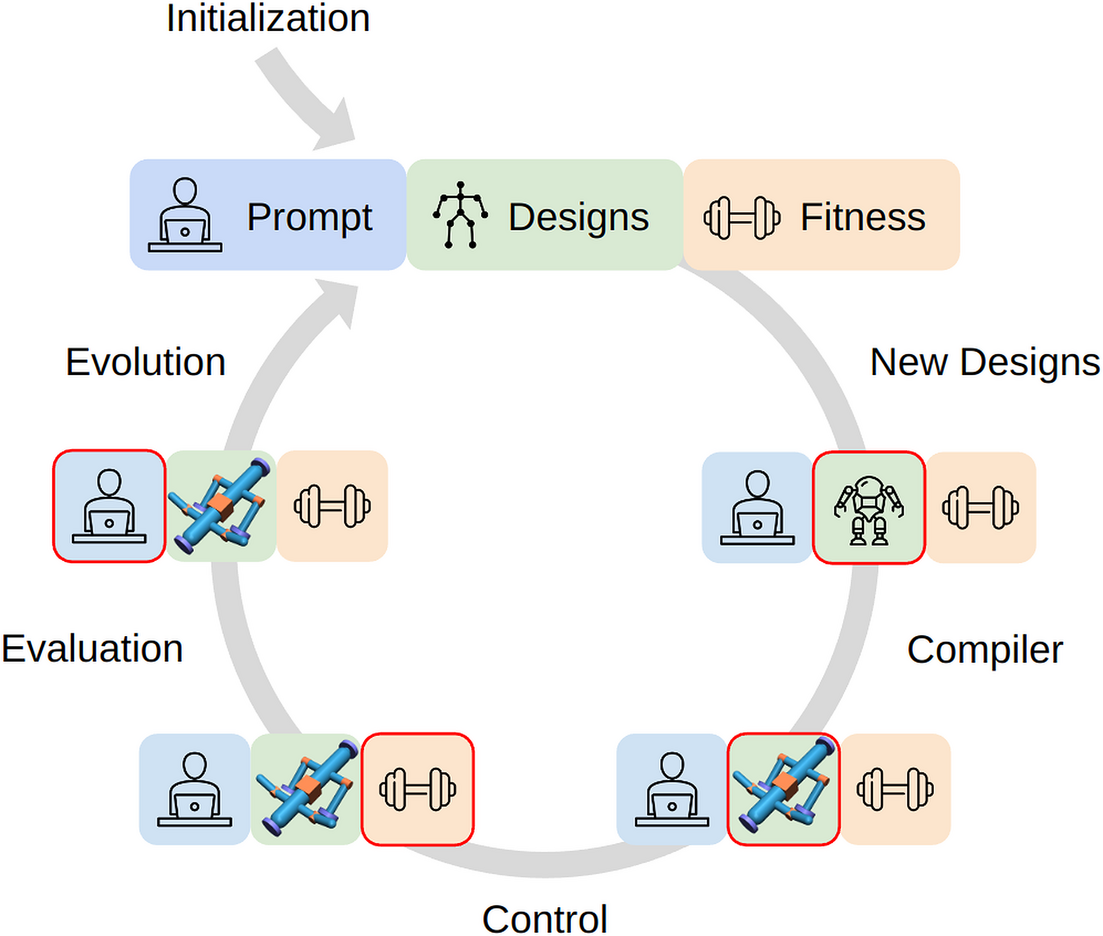}
    \caption{One iteration of \method{}. At each stage, the \textcolor{blue}{prompt}, \textcolor{green!60!black}{robot design}, and \textcolor{orange}{fitness score} are displayed. The element modified in each step is highlighted in red. This closed-loop cycle enables progressive design refinement through evolutionary selection.}
    \vspace{-1em}
    \label{fig:framework_loop}
\end{figure}

The grammar, adapted from Robogrammar~\cite{zhao2020robogrammar}, defines structural rules that ensure generated designs are physically plausible. At each generation, the LLM proposes new robot designs expressed in this grammar. These are compiled into XML models and simulated to train control policies, which determine each design's fitness score. The highest-performing designs are retained and incorporated as few-shot exemplars in the next prompt, enabling iterative improvement. Figure~\ref{fig:framework_loop} provides an overview of this process.

\begin{figure*}[t]
\centering
\includegraphics[width=\textwidth]{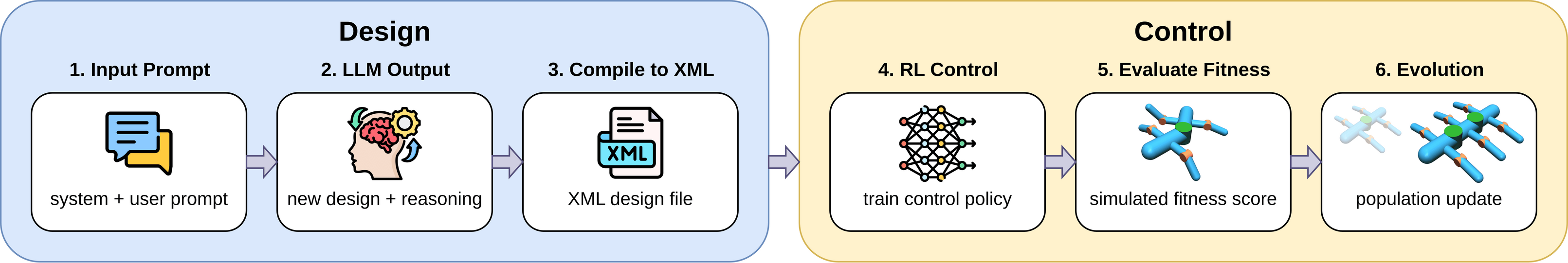}
\caption{Overview of the \method{} framework, which consists of a \textit{design stage} and a \textit{control stage}. The process is divided into six steps:
(1) The input consists of the system prompt, user prompt, and few-shot examples.
(2) Based on the input prompt, the LLM generates a new design and provides its reasoning.
(3) A compiler converts the LLM-generated design into an XML file.
(4) An RL control policy is trained in simulation for the robot in a given environment.
(5) The robot is evaluated, and a fitness score is assigned to the design.
(6) The population of robots is updated through evolution by pruning the worst-performing candidates and feeding the top designs back as examples for the next iteration. Together, these stages form an automated design loop that requires no human intervention after initialization.}
\label{fig:robomorph}
\end{figure*}

LLMs complement evolutionary algorithms in three ways. First, they explore the morphology space more efficiently than hand-coded mutation operators. Second, they provide a natural-language interface for specifying design constraints. Third, they generate structurally diverse candidates in a single step rather than through incremental perturbations.

While our experiments are conducted in simulation, manufacturing constraints and sim-to-real transfer remain open challenges. Nevertheless, the results demonstrate the feasibility of using LLMs as generative operators for morphology optimization. The designs discovered by \method{} exhibit distinctive traits that reflect the LLM's capacity to recombine structural rules in unexpected ways.

In summary, our contributions are as follows:
\begin{itemize}[noitemsep, topsep=0pt]
    \item We introduce \method{}, the first framework to integrate LLM-based generation, evolutionary algorithms, robot grammars, and RL control for automated robot morphology design.
    \item We demonstrate that treating the LLM as a mutation operator yields competitive or superior designs compared to Robogrammar’s graph-search method, while producing more structurally diverse morphologies.
    \item We analyze design outcomes across four terrains, highlight the strengths and limitations of our approach, and outline directions for bridging simulation with real-world deployment.
\end{itemize}

%% file: sections/method.tex
\section{\method{}}
\label{sec:method}

The \method{} framework, shown in Figure~\ref{fig:robomorph}, follows an evolutionary pipeline that maintains a population of robot designs generated by an LLM (specifically, GPT-4o). At each iteration, the LLM is prompted with few-shot examples and produces new designs expressed in a grammar-based representation~\cite{zhao2020robogrammar} (Sections~\ref{sec:prompt} and~\ref{sec:few_shots}). The grammar enables diverse morphologies while ensuring outputs are physically valid and fabricable.

Each design is compiled into an XML file (Section~\ref{sec:robot_design}) and simulated in MuJoCo~\cite{todorov2012mujoco} via MJX (Section~\ref{sec:simulation}) to train a control policy and compute a fitness score (Section~\ref{sec:control}). The population is updated by retaining the top-$K$ designs, which also serve as few-shot examples for the next iteration (Section~\ref{sec:evolution}). The process is initialized with randomly generated examples and summarized in Algorithm~\ref{alg:method}.

\begin{algorithm}[ht]
    \caption{Pseudocode for \method{}} 
    \label{alg:method}
    \begin{algorithmic}\small
        \STATE \textbf{Requires:}
        \STATE $\mathtt{prompt_{system}}$ \hfill robot grammar rules  
        \STATE $\mathtt{prompt_{user}}$ \hfill instructions for designing a robot  
        \STATE $\mathtt{evolutions}$ \hfill number of evolutions  
        \STATE $\mathtt{population}$ \hfill number of designs per population  
        \STATE $K$ \hfill number of few-shot examples  
        \STATE $\mathcal{Q}$ \hfill min-queue $\{(\mathtt{fitness}_i, \mathtt{design}_i, \mathtt{reason}_i)\}_{i=1}^{K}$
        \STATE

        \STATE \textbf{function} \method{}
        \STATE $\mathtt{fewshots} \gets$ generate $K$ random designs by sampling grammar rules
        \FOR{$i = 1$ \textbf{to} $\mathtt{evolutions}$}
            \FOR{$j = 1$ \textbf{to} $\mathtt{population}$}
                \STATE $\mathtt{design_{new}, reason_{new}} \gets LLM(\mathtt{prompt_{system}}, \mathtt{prompt_{user}}, \mathtt{fewshots})$
                \STATE $\mathtt{design_{xml}} \gets Compiler(\mathtt{design_{new}})$
                \STATE \textbf{if} $\mathtt{design_{xml}}$ is corrupted \textbf{then continue} \hfill \textit{// skip to next $j$}
                \STATE $\mathtt{fitness_{new}} = Evaluate(\mathtt{design_{xml}})$
                \STATE $\gamma_{\min} \gets \arg\min_{q} \mathtt{fitness}_q$ in $\mathcal{Q}$
                \IF{$\mathtt{fitness_{new}} > \mathcal{Q}[\gamma_{\min}].\mathtt{fitness}$}
                    \STATE $\mathcal{Q}[\gamma_{\min}] \gets (\mathtt{fitness_{new}}, \mathtt{design_{new}}, \mathtt{reason_{new}})$
                \ENDIF
            \ENDFOR
            \STATE $\mathtt{fewshots} \gets \emptyset$
            \FORALL{$(\mathtt{fitness}, \mathtt{design}, \mathtt{reason}) \in \mathcal{Q}$}
                \STATE $\mathtt{fewshots.append}(\textbf{str}(\mathtt{design}, \mathtt{reason}))$
            \ENDFOR
        \ENDFOR
        \STATE \textbf{end function}
    \end{algorithmic}
\end{algorithm}

\subsection{Prompt Structure}\label{sec:prompt}
We generate robot designs using GPT-4o with three main prompt components:

\subsubsection{System prompt} Specifies the robot grammar rules~\cite{zhao2020robogrammar}, which constrain outputs to physically realizable designs. In effect, the system prompt describes the ``building blocks'' and rules for assembling them. The LLM is instructed to apply a reasoning trace detailing a step-by-step application of these rules during the assembly process. This not only clarifies the LLM's design rationale but also helps the model avoid violating the grammar constraints.

\subsubsection{User prompt} Provides a fixed high-level instruction:  
\texttt{``Your task is to design a single robot. Use the examples provided to guide your reasoning and explain your design step by step.''}  
This chain-of-thought style~\cite{wei2022chain} encourages sequential reasoning and improves the interpretability of the LLM's outputs.

\subsubsection{Few-shot examples} Include top-performing designs (with reasoning traces) from previous iterations. These serve as templates, allowing the LLM to reuse effective traits while maintaining diversity. We omit explicit fitness scores, as providing them caused the model to overfit and repeatedly propose similar designs, thereby reducing exploration.

\subsection{Best-Shot Prompting}
\label{sec:few_shots}
Our prompting strategy, inspired by in-context learning~\cite{brown2020language}, augments the prompt with high-quality few-shot examples. We call this approach \textit{best-shot prompting}, motivated by three factors:

\textbf{Reducing invalid outputs.} In zero-shot trials, the LLM often produced designs that violated grammar rules, yielding XML files that could not be compiled. Providing valid designs with reasoning steps grounds the model in the grammar and significantly reduces such errors~\cite{romera2024mathematical}. In our experiments, most zero-shot outputs failed to compile, whereas with best-shot prompting and chain-of-thought reasoning, the vast majority of designs were grammar-valid.

\begin{figure}[ht]
\centering
\includegraphics[width=\columnwidth]{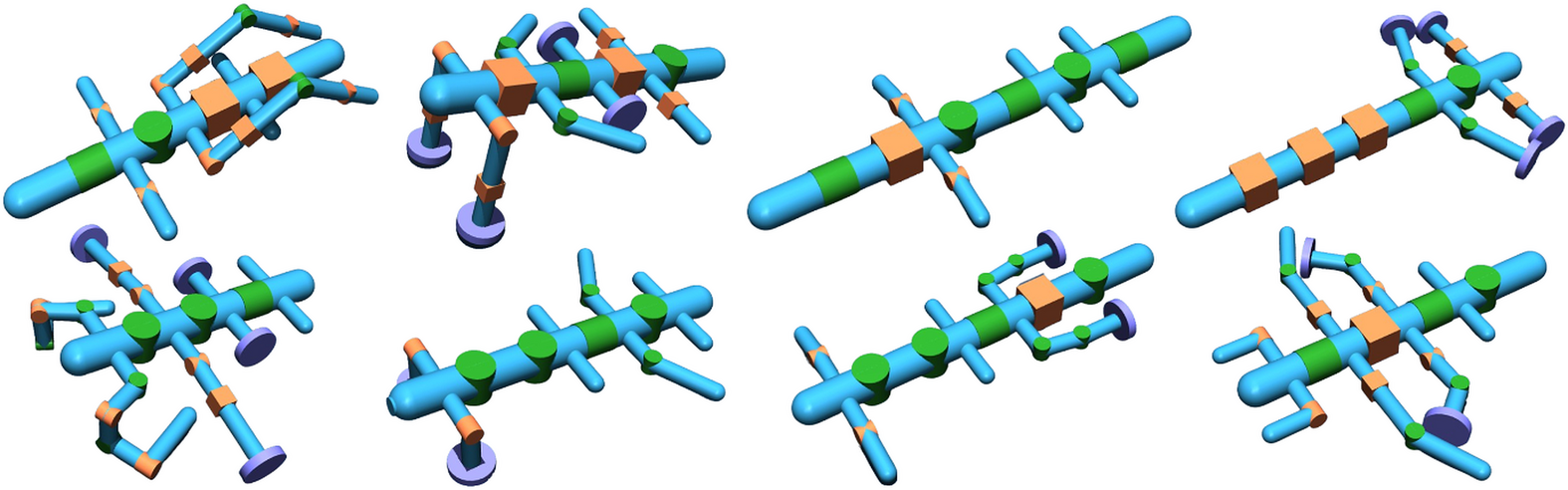}
\caption{Examples of randomly generated initial few-shot robot designs used to seed Algorithm~\ref{alg:method}. These serve as diverse seed morphologies to prime the evolutionary process.}
\vspace{-1em}
\label{fig:init_fewshots}
\end{figure}

\textbf{Encouraging diversity.} Without diverse exemplars, the LLM tended to repeat small variations of one concept. We seed prompts with randomly generated designs (Figure~\ref{fig:init_fewshots}) to span different structures, and update them with top performers each generation. This maintains variety in limb count, joint types, and symmetries, preventing premature convergence.

\textbf{Iterative improvement.} Updating few-shot examples with the current best designs introduces an implicit elite-selection effect. The prompt gradually shifts toward stronger exemplars, guiding the LLM to propose increasingly effective morphologies. We exclude numeric fitness values because including them led to overfitting and reduced exploration. Instead, the model infers successful traits from qualitative patterns in the examples, often producing unconventional designs that are then filtered by evolution (Section~\ref{sec:evolution}).

\subsection{Robot Design}
\label{sec:robot_design}
Our design space is defined by a context-free grammar adapted from Robogrammar~\cite{zhao2020robogrammar}, which specifies how robots can be constructed from a small set of modules. The grammar enforces bilateral symmetry, meaning any limb added to one side of the body must be mirrored on the opposite side. This constraint simplifies locomotion, but also reduces the effective diversity of possible morphologies. Each expansion rule in the grammar maps directly to a physically realizable module.

As shown in Robogrammar, such a grammar can describe hundreds of thousands of designs while guaranteeing that every generated robot can be assembled from modular parts. In our framework, this ensures that the LLM’s proposals always correspond to feasible designs, though the space is limited by the chosen module library.

The LLM generates textual representations of a robot design in the grammar formalism, which avoids the need for explicit tree search as in Robogrammar. This significantly accelerates design generation, as the LLM proposes complete structures directly. A compiler converts the text into an XML description compatible with MuJoCo (via MJX). If the design violates the grammar, the compiler fails to produce a valid XML and the output is discarded. In practice, invalid outputs are uncommon, and most designs compile successfully once the few-shot examples are in place. 

After obtaining a valid XML model of the robot, we disable self-collisions for the duration of policy training to accelerate simulation, a trade-off also adopted in prior work~\cite{gupta2021embodied}. This simplification may allow physically unrealistic interpenetration. However, our tree-structured grammars produce mostly open morphologies where self-collisions are infrequent, and we apply this condition uniformly across all designs. In Sections~\ref{sec:results} and~\ref{sec:benchmark}, we re-enable collisions for final evaluation. The top designs retained their effectiveness under full collision physics, confirming physical plausibility.

\subsection{Simulation}
\label{sec:simulation}
A key challenge in our approach is that each new robot design must be evaluated by training a controller in simulation, which is computationally expensive. We use MuJoCo XLA (MJX), a JAX-based implementation of MuJoCo~\cite{todorov2012mujoco} capable of running on accelerated hardware, combined with Brax~\cite{brax2021github} for RL algorithms such as SAC~\cite{haarnoja2018soft} and PPO~\cite{schulman2017ppo}. MJX uses GPUs to parallelize experience collection across thousands of environment instances. We further parallelize RL training across multiple GPUs, allowing several designs to be trained simultaneously. This combined speedup makes the evolutionary loop practical by significantly increasing both the number of iterations and designs evaluated per generation.

\subsection{Control}
\label{sec:control}
For each valid design, we train a locomotion controller in simulation using the SAC algorithm, and the resulting policy is evaluated to compute a fitness score. We also experimented with PPO, but found SAC to be more consistent across a diverse range of robot morphologies. We did not tailor the RL parameters to each morphology or terrain, as each design was evaluated using the same set of hyperparameters and for the same number of environment steps.

\subsection{Evolution}
\label{sec:evolution}

We treat the LLM as a stochastic design generator analogous to a mutation operator. In each generation, we sample the LLM multiple times to obtain a diverse set of candidate designs. A grammar verifier filters out any invalid outputs, ensuring that only structurally sound candidates enter the evaluation pipeline. This combination of LLM stochasticity and grammar-based verification balances broad exploration with physical plausibility.

After all designs in a generation have been evaluated, we rank them by fitness and retain the top-$K$ candidates, favoring the ``survival'' of better designs. This introduces an elite-selection effect, and the top-performing designs form the few-shot examples for the LLM in the next generation. Over successive generations, this evolutionary loop of variation and selection improves the population's overall fitness. Three mechanisms maintain exploration during exploitation. First, LLM sampling stochasticity ensures each call produces a distinct design. Second, excluding fitness scores from prompts prevents overfitting and preserves structural variety (Section~\ref{sec:ablation}). Third, randomly generated initial examples establish a broad starting distribution.

%% file: sections/results.tex
\section{Experiments}
\label{sec:experiments}

\subsection{Experimental Setup}\label{sec:experimental_setup}
We evaluate \method{} using 8 random seeds over 50 evolutionary generations, with a population size of 32 designs per evolution, $K{=}3$ few-shot examples, and GPT-4o at its default temperature of 1.0. Given the stochastic nature of the LLM's output, our initial grid search over these hyperparameters proved ineffective, as each combination yielded varying results across seeds. We therefore chose reasonable values for these hyperparameters based on similar work in~\cite{ringel2024text2robot}. For completeness, we also conduct ablation studies on (i) LLM variants and (ii) evolution strategy, which are presented in Section~\ref{sec:ablation}. In total, 1,600 robot designs are evaluated per seed. During RL policy training, we evaluate 8 robot designs simultaneously in a computing cluster equipped with 8 NVIDIA A100-40GB GPUs. The total runtime of \method{} for a single seed amounts to approximately 200 GPU hours. We provide further details on the computational time of our framework in Section~\ref{sec:computation}.

We assess the fitness of each design within a custom MJX environment and use the Brax implementation of the SAC algorithm. The system state is represented by a vector comprising the position and velocity of each joint, while the action space consists of continuous joint torques in the range $(-1, 1)$. The per-timestep reward is defined in Equation~\ref{eq:reward} as the instantaneous forward velocity of the robot's head (the most forward body part) as it transitions from state $s$ to $s'$ under action $a$:
\begin{equation}
r_t(s,a,s') = \vec{v_x}(s,a,s'),
\label{eq:reward}
\end{equation}
where $\vec{v_x}$ denotes the robot head’s center-of-mass velocity in the $x$ (forward) direction. Rollouts are terminated early if the agent becomes unhealthy (based on $z$-height) or when the episode length is reached.

The overall fitness score of a design, given in Equation~\ref{eq:overall_fitness}, is defined as the average reward achieved by its learned policy. We estimate this via Monte Carlo simulation across $N=128$ rollouts:
\begin{equation}
F(\mathtt{design}) = \frac{1}{N} \sum_{i = 1}^{N} \frac{1}{T_i} {\sum_{j = 1}^{T_i}{r_{ij}}},
\label{eq:overall_fitness}
\end{equation}
where $T_i$ is the length of the $i$-th rollout and $r_{ij}$ is the reward obtained at the $j$-th step of that rollout.

For policy learning, both the critic and policy networks are 2-layer MLPs with 256 units per layer. We employ the Adam optimizer~\cite{kingma2014adam} and use the ReLU activation function across all layers in each network. Each policy is trained for $10^6$ environment steps with a batch size of 256, which we found provides a good balance between computational efficiency and training stability. All other RL-related hyperparameters are kept constant across all experiments for fairness and are listed in our repository.

We evaluate designs on four terrain environments that present distinct locomotion challenges. These are adapted from Robogrammar, with modifications where simulator limitations prevented exact replication. Together, they test robustness across smooth, uneven, low-friction, and constrained settings. Figure~\ref{fig:designs} provides visualizations of each terrain.

\textbf{Ridged terrain.} A series of periodic ridges that require climbing or crawling strategies for effective navigation.  

\textbf{Flat terrain.} A featureless, high-friction plane suitable for a broad range of locomotion modes.  

\textbf{Frozen terrain.} A low-friction surface that demands stability and efficient traction for forward movement.

\textbf{Beams terrain.} A series of platforms with overhead beams that penalize tall or upright morphologies and reward compact, low-profile designs.  

\subsection{Main Experiment}\label{sec:results}
\begin{figure}[ht]
    \centering
    \begin{subfigure}[t]{\linewidth}
        \centering
        \includegraphics[width=0.9\linewidth]{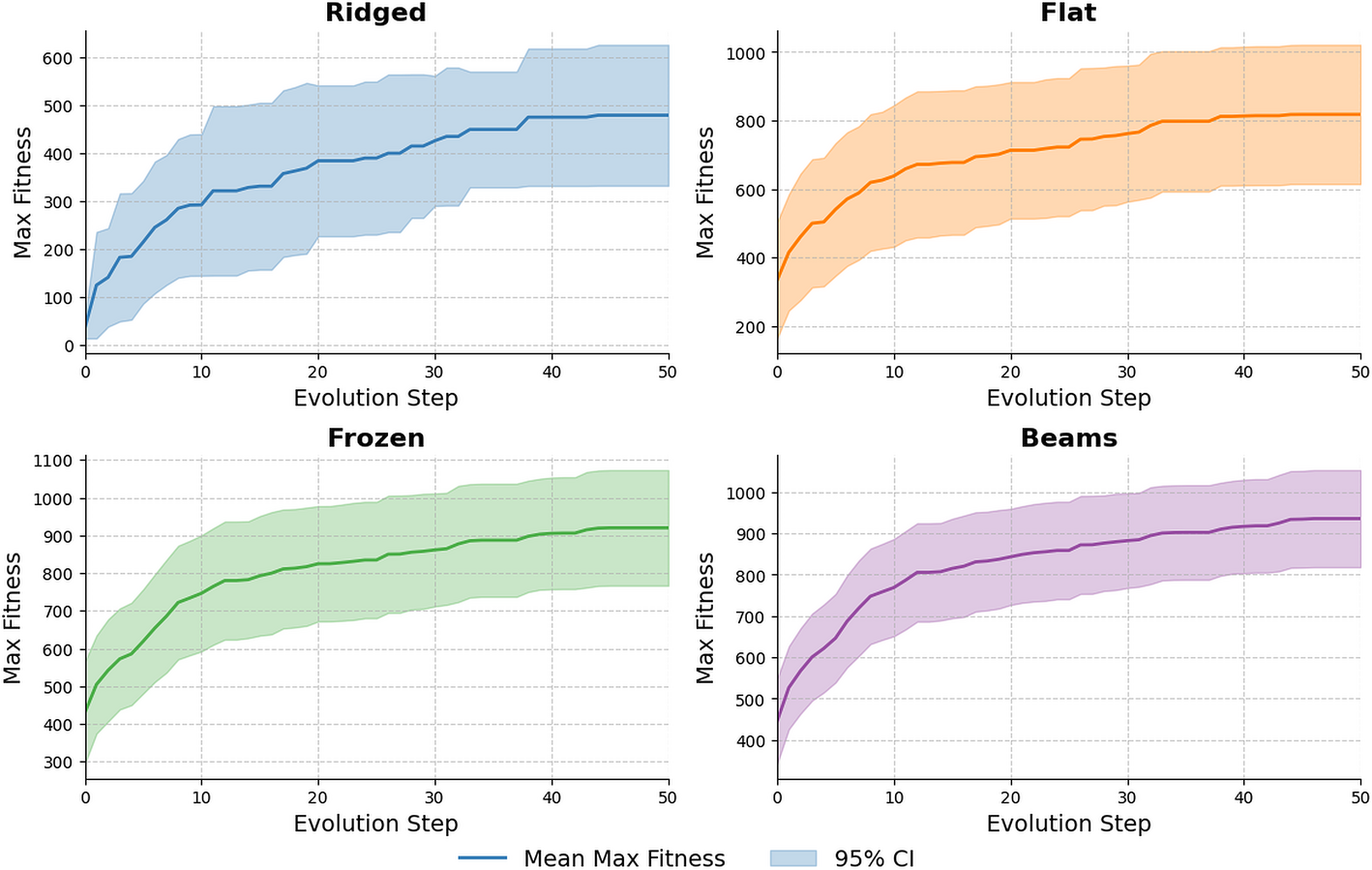}
        \caption{Average maximum fitness with 95\% confidence intervals across 8 seeds for the best robot design in the population at each evolutionary step.}
        \label{fig:terrain_plot}
    \end{subfigure}
    \begin{subfigure}[t]{\linewidth}
        \centering
        \includegraphics[width=0.9\linewidth]{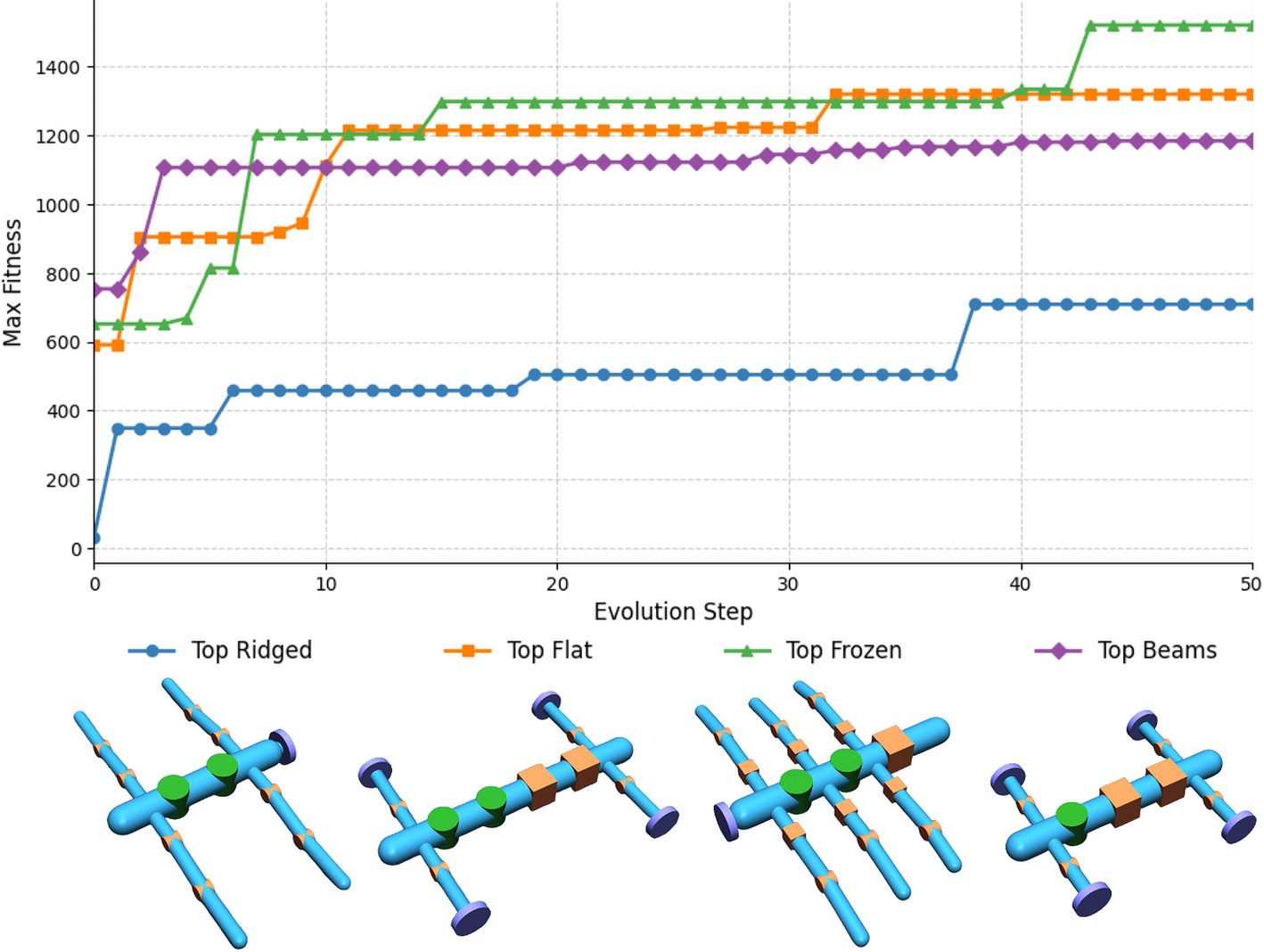}
        \caption{Evolution of the best-performing seed for each terrain.}
        \label{fig:best_seed}
    \end{subfigure}
    \caption{Maximum fitness of the best robot design in the population at each evolutionary step. Fitness consistently improves across generations, confirming that \method{} progressively improves design quality.}
    \label{fig:main_results}
\end{figure}

We first evaluate \method{} over the course of evolution. Figure~\ref{fig:terrain_plot} shows the average of the maximum fitness scores for the top design in the population at each evolutionary generation, across 8 seeds (shaded region denotes 95\% confidence). A positive trend indicates that successive iterations of \method{} produce progressively better designs. The maximum fitness is monotonically non-decreasing by construction due to elite retention across generations.

\begin{figure}[ht]
    \centering
    \includegraphics[width=0.8\columnwidth]{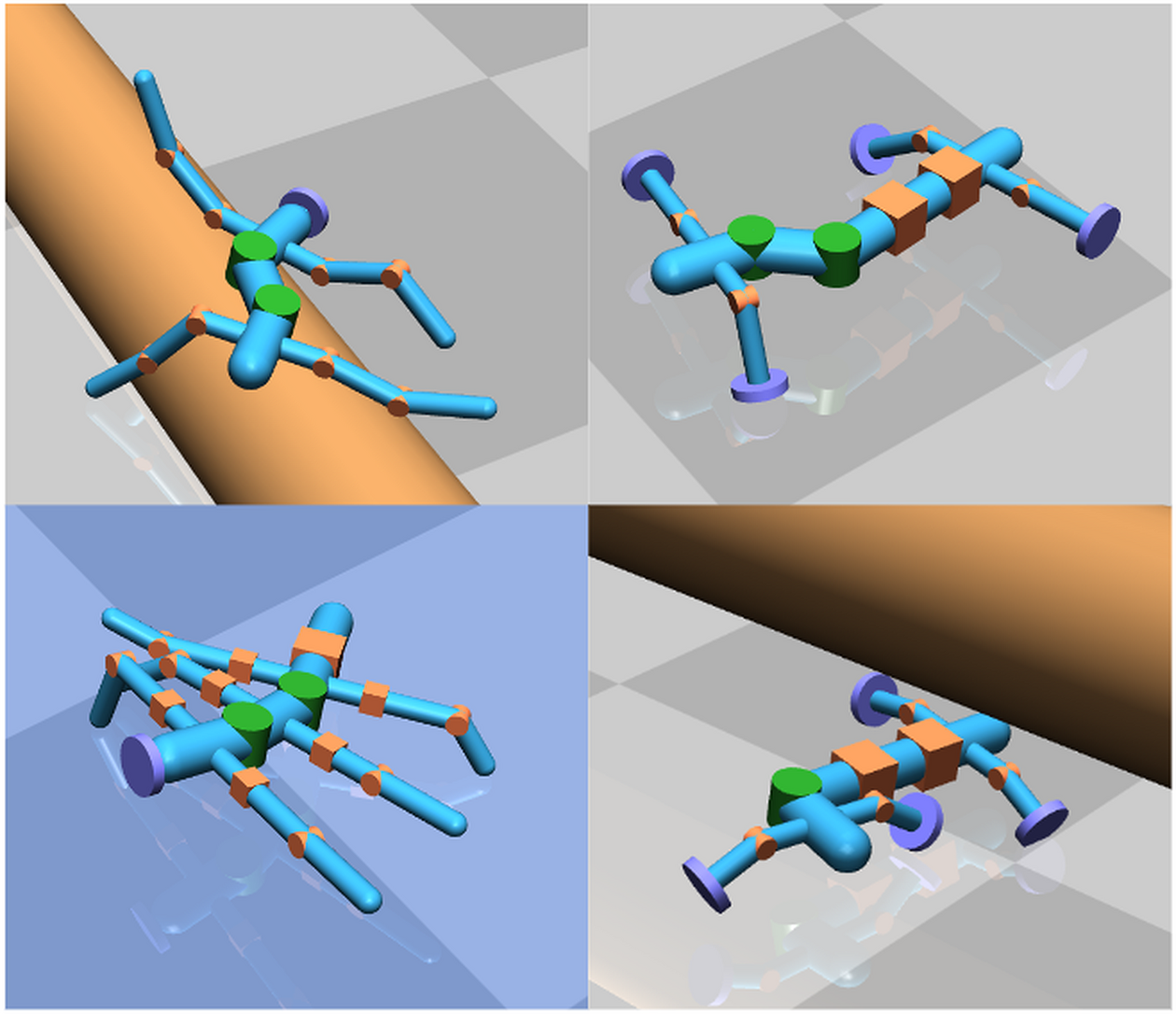}
    \caption{Top robot designs from \method{} per terrain: ridged (top-left), flat (top-right), frozen (bottom-left), and beams (bottom-right). Each terrain yields a distinct morphology, including wheeled and hexapod designs not found by prior methods.}
    \label{fig:designs}
    \vspace{-1em}
\end{figure}

Figure~\ref{fig:best_seed} depicts the evolution of the best-performing seed on each terrain, along with the highest overall fitness attained. This represents the exploitation phase of our evolutionary algorithm (see Section~\ref{sec:evolution}), showing the best result \method{} discovered for each environment. The top-performing designs for each of the four terrains are visualized in Figure~\ref{fig:designs}. Each terrain yields a distinct morphology tailored to the challenges of that environment.

\textbf{Ridged terrain.} A quadruped with elongated limbs and knee joints that can swing upward, allowing the robot to step over ridges and clear obstacles with high foot clearance.

\textbf{Flat terrain.} A low-profile body with shorter limbs ending in unactuated wheels spaced far apart. This wheeled quadruped exploits continuous rolling contact with the ground, minimizing energy loss and enabling faster locomotion than its legged counterpart. The low inertia of the short limbs, combined with free-spinning wheels, allows for smoother movements on flat surfaces.

\textbf{Frozen terrain.} A hexapod with rigid upper-leg joints that keep the legs splayed outward, creating a wide support base and low center of gravity. This configuration provides stability on slippery surfaces by resisting lateral sliding. Hexapods inherently offer greater static stability by always having at least three legs in contact with the ground, keeping the center of mass within the support polygon~\cite{mcghee2007adaptive}.

\textbf{Beams terrain.} Morphologically similar to the flat-terrain design, but with a lower profile and shorter limbs to maintain clearance under overhead beams. This similarity validates the framework, as identical ground conditions yield similar strategies, with adaptation only where the environment demands it (the overhead height constraint).

\subsection{\method{} vs. Robogrammar}\label{sec:benchmark}
\begin{figure}[ht]
    \centering
    \begin{subfigure}[t]{0.49\linewidth}
        \centering
        \includegraphics[width=\linewidth]{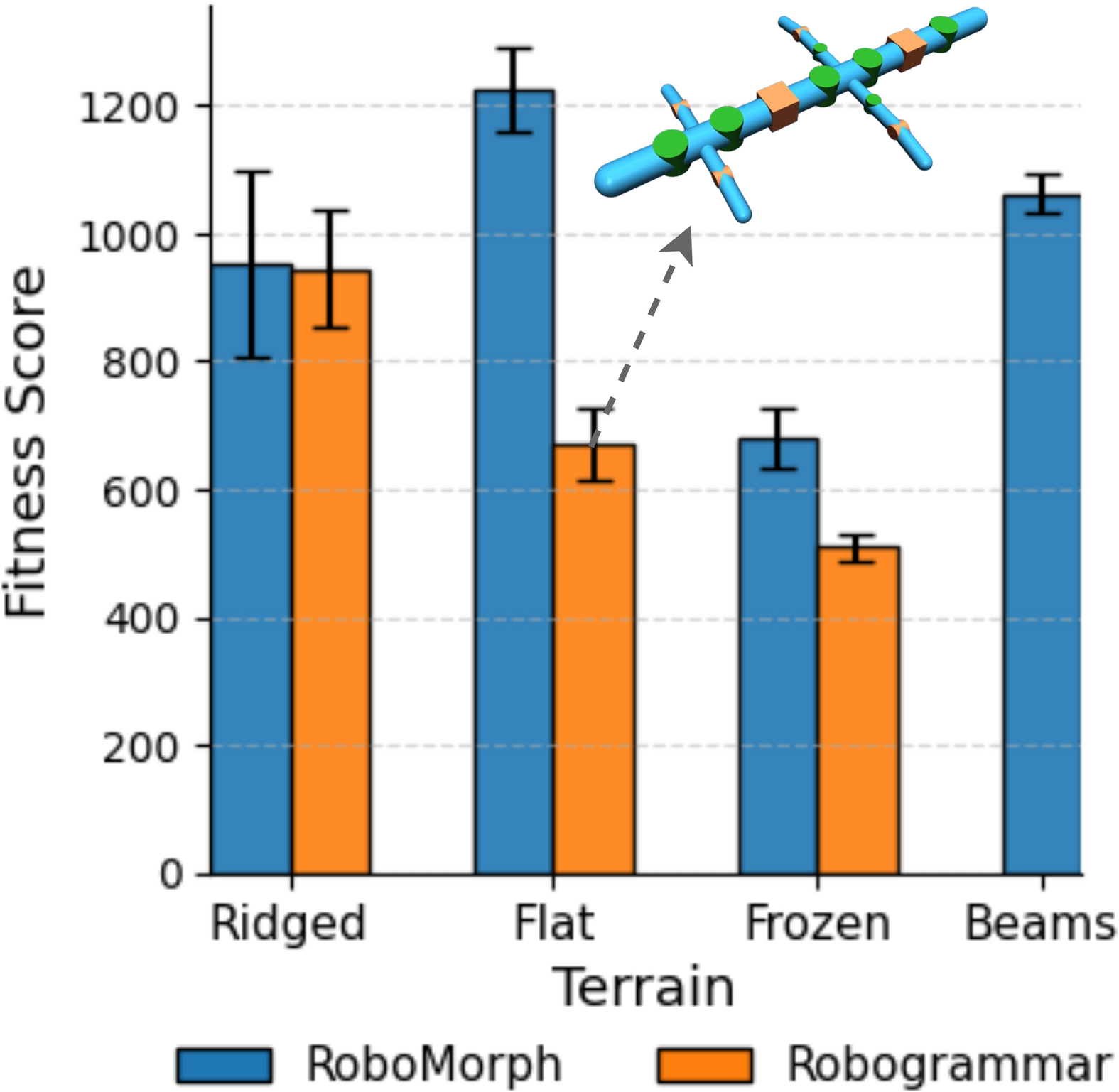}
        \caption{Evaluation using RL.}
        \label{fig:rl_eval}
    \end{subfigure}
    \hfill
    \begin{subfigure}[t]{0.49\linewidth}
        \centering
        \includegraphics[width=\linewidth]{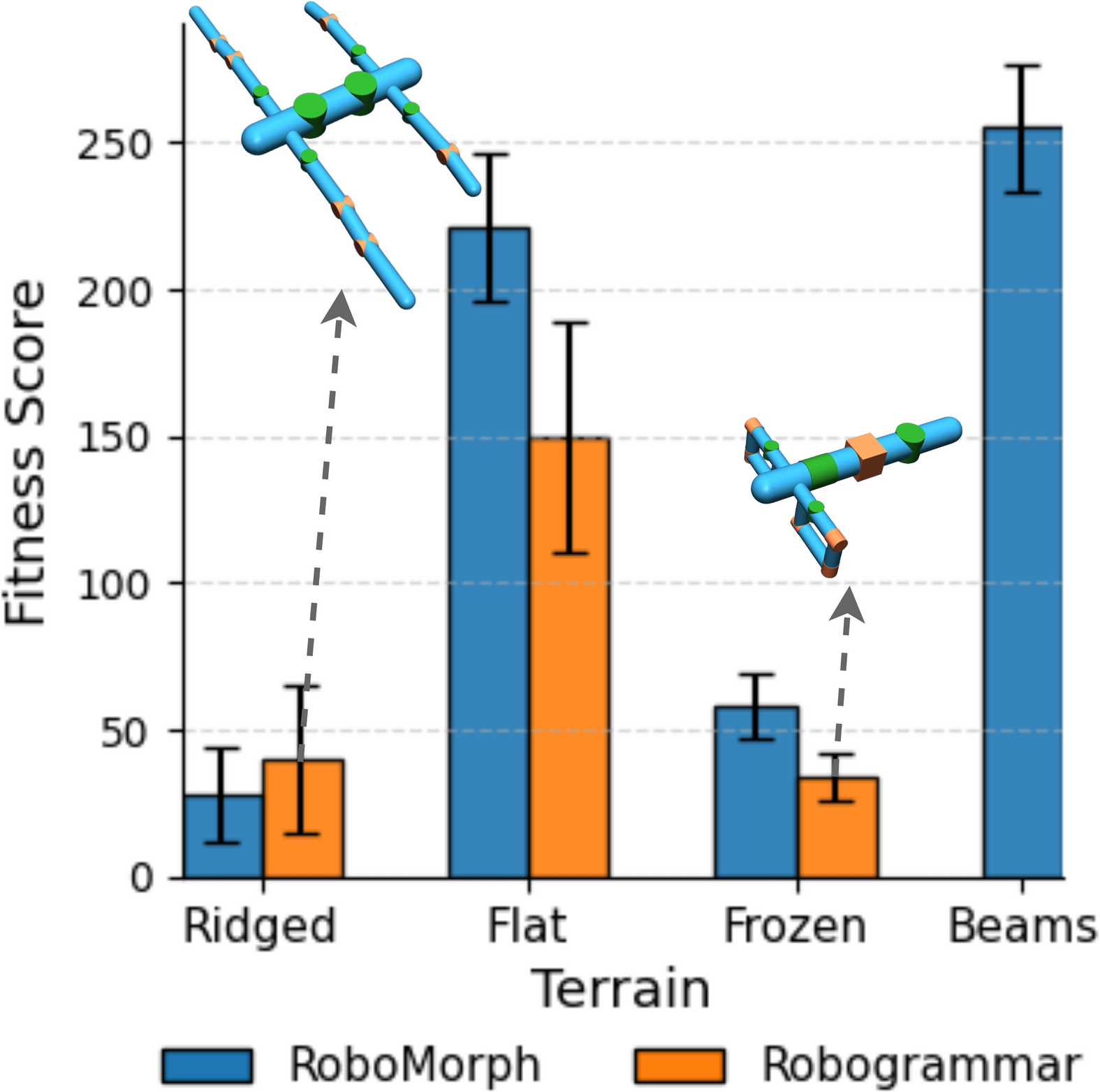}
        \caption{Evaluation using MPC.}
        \label{fig:mpc_eval}
    \end{subfigure}
    \caption{Average fitness of top designs from RoboMorph and Robogrammar across 8 seeds per terrain. Under RL, \method{} matches or outperforms Robogrammar on all shared terrains. Under MPC, results are comparable on ridged terrain, while \method{} outperforms on flat and frozen.}
    \label{fig:evaluation}
\end{figure}

We compare \method{} against Robogrammar by reconstructing their best-performing designs for ridged, flat, and frozen terrains in our MJX environment (Figure~\ref{fig:evaluation}). Robogrammar's gapped terrain could not be replicated due to MJX collision geometry limitations, and our beams terrain has no Robogrammar baseline.

For a fair comparison, we re-enabled full collisions for both methods' designs (disabled during evolutionary search to reduce computation, as noted in Section~\ref{sec:robot_design}). We train an RL policy from scratch for each design and repeat training 8 times per design to account for variability\footnote{We take the highest-fitness design found by \method{} across all evolution runs and evaluate it with 8 independent RL trainings.}. Results under RL control are shown in Figure~\ref{fig:rl_eval}.

Since Robogrammar used a model-predictive control (MPC) planner, we also compare under MPC by replicating their MPPI-based controller (Figure~\ref{fig:mpc_eval}). This comparison is inherently asymmetric, as \method{}'s designs were evolved under RL-based fitness while Robogrammar's were optimized with MPC. Evaluating under both controllers mitigates this bias, and consistent performance trends across both schemes support the validity of our conclusions.

Figure~\ref{fig:evaluation} compares performance of both methods on each terrain. On ridged terrain, both methods achieve similar fitness scores with overlapping error bars. This result is expected given that the two designs are morphologically similar. Both are quadrupeds featuring roll joints in the body and a comparable arrangement of leg joints. On flat terrain, \method{} significantly outperforms Robogrammar's leg-only solution, as the discovered wheeled design yields faster and more efficient locomotion. Robogrammar's graph search does not readily explore wheeled configurations because incremental structural mutations are unlikely to produce the specific joint and wheel combination that enables rolling contact.

On frozen terrain, \method{} also produces a higher-performing design than Robogrammar's bipedal structure, which relies on compact arms for ground contact while the rear body slides freely. Our approach instead discovered a hexapod with a wide base and continuous contact points, which is more efficient at generating forward thrust on a slippery surface. Overall, the results highlight that \method{} discovers novel, effective strategies that Robogrammar's search did not find.

\subsection{Ablation Studies}\label{sec:ablation}
\begin{figure}
    \centering
    \begin{subfigure}[t]{0.49\linewidth}
        \centering
        \includegraphics[width=\linewidth]{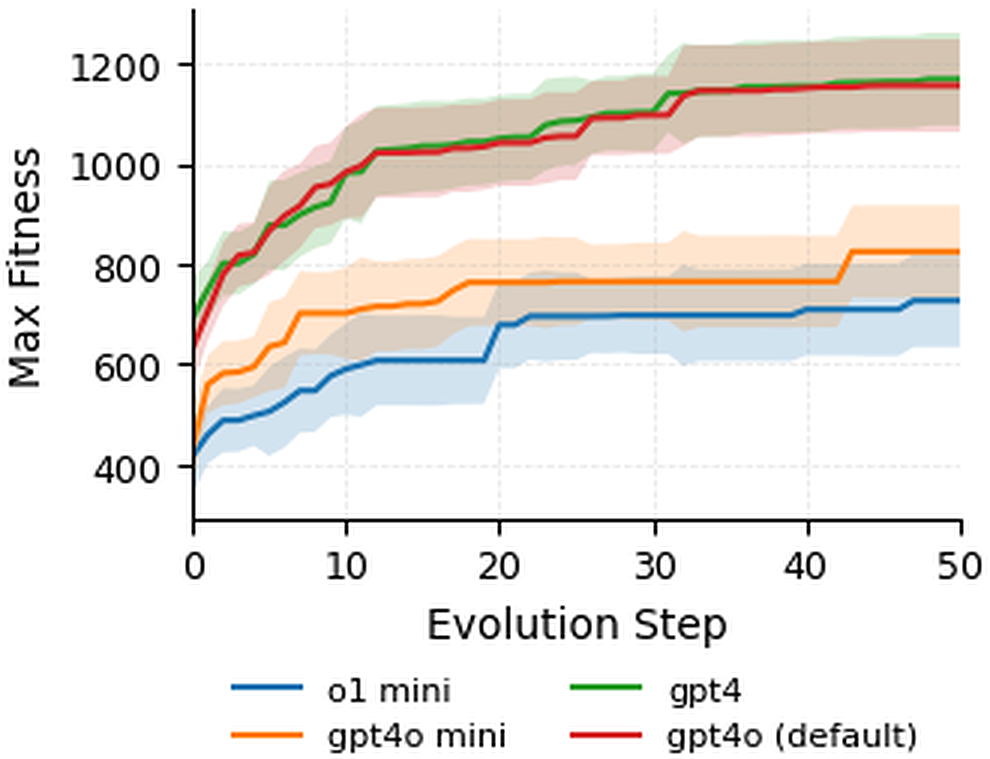}
        \caption{LLM variant ablation.}
        \label{fig:ablation_llm}
    \end{subfigure}
    \hfill
    \begin{subfigure}[t]{0.49\linewidth}
        \centering
        \includegraphics[width=\linewidth]{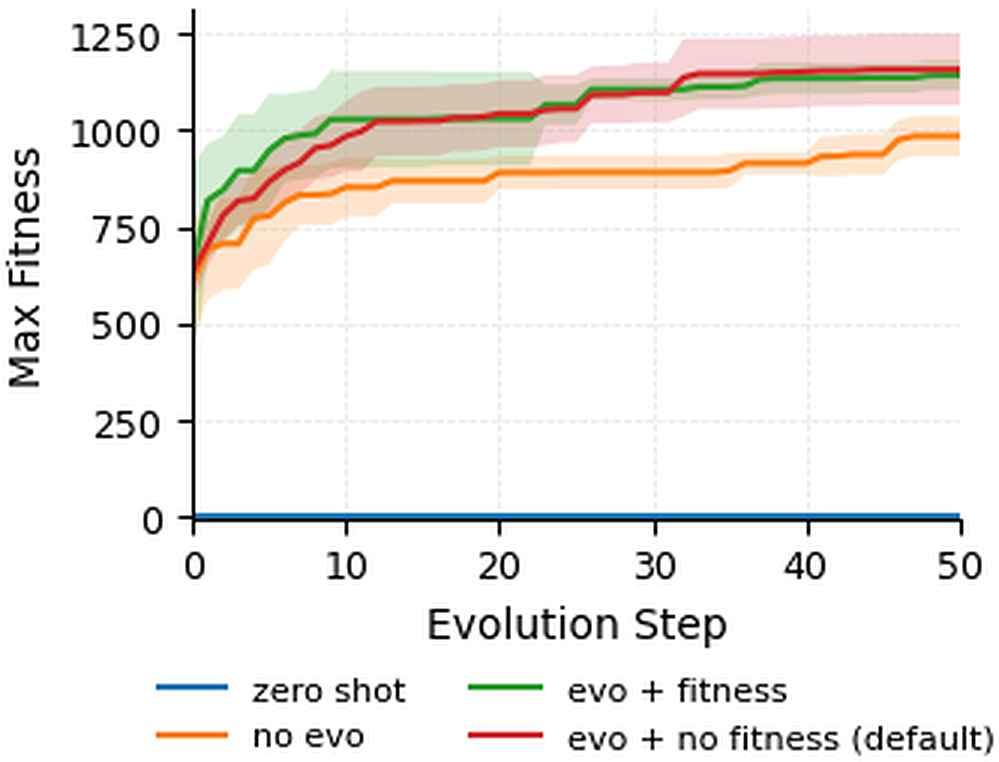}
        \caption{Evolution ablation.}
        \label{fig:ablation_evo}
    \end{subfigure}
    \caption{Ablation on the flat terrain: average maximum fitness with 95\% confidence intervals across 8 seeds. Larger LLMs and evolutionary prompting both yield higher fitness.}
    \label{fig:ablation}
\end{figure}

We conduct targeted ablation studies to isolate the impact of two main design choices in our pipeline: the LLM model variant and evolutionary prompting strategy. The results are summarized in Figure~\ref{fig:ablation}.

\subsubsection{LLM variant} We compared several LLM back-ends for the design generation module, including GPT-4o (default), GPT-4o mini, GPT-4, and o1-mini. As seen in Figure~\ref{fig:ablation_llm}, both GPT-4o and GPT-4 produce significantly higher-fitness designs than the smaller variants. GPT-4 yielded only marginal improvements over GPT-4o but at substantially higher cost (\$30/\$60 per million input/output tokens vs. \$2.50/\$10 for GPT-4o).

\subsubsection{Evolution strategy} We compared four evolutionary strategies to understand the contribution of each component. We tested: (i) zero-shot baseline with no optimization, (ii) prompting without any evolution, (iii) evolution with fitness scores, and (iv) evolution without fitness scores. Results in Figure~\ref{fig:ablation_evo} show that incorporating evolution in the pipeline outperforms non-evolution baselines, confirming that iterative refinement is key. Explicit fitness scores provide only minor benefit and can reduce diversity by overfitting to top designs. Excluding scores maintains exploration while still improving performance, supporting our choice in Section~\ref{sec:few_shots}.

\subsection{Computational Time}\label{sec:computation}
\textbf{Overall compute.} \method{} requires approximately 200 GPU hours per seed on A100 GPUs, compared to Robogrammar's reported 31 CPU hours~\cite{zhao2020robogrammar}. In total compute, \method{} is substantially more expensive, and these numbers correspond to different hardware classes (GPUs vs. CPUs), making direct comparison difficult. However, our compute is dominated by RL training, which is highly parallelizable. On 8 A100 GPUs, a full run completes in about 25 wall-clock hours. While Robogrammar invests most of its compute in sequential graph search, \method{} shifts effort to parallelizable RL evaluation. This parallelism allows \method{} to converge within about a day, making it practical for iterative development despite higher total resource usage.

\textbf{Design generation.} Each GPT-4o API call takes approximately 9.5s (1s time-to-first-token plus $\sim$1500 output tokens at 175 tokens/s)~\cite{gpt4o-providers}. For one seed with 1600 evaluations, cumulative design generation time is $\sim$4.2 hours, confirming that RL evaluation dominates total compute. By comparison, Robogrammar allocates 11 of its 31 total hours to design generation.

\subsection{Discussion}\label{sec:discussion}

\textbf{On the LLM's contribution.} Our ablation studies (Section~\ref{sec:ablation}) demonstrate that iterative evolution with best-shot prompting yields consistent improvements over zero-shot and non-evolutionary baselines. However, we acknowledge that our experiments do not include a traditional evolutionary algorithm (e.g., random mutation and crossover on the same grammar) as a baseline. Such a comparison would help isolate whether the LLM's structured generation provides an advantage beyond acting as a stochastic design sampler. Unlike random mutation, the LLM produces complete, grammar-valid designs with reasoning traces, which helps explain its effectiveness in navigating the combinatorial design space. Crucially, the LLM can make non-local jumps in design space, proposing an entirely different body plan (e.g., shifting from a quadruped to a hexapod) in a single generation, whereas traditional mutation operators are limited to incremental structural edits. Recent work~\cite{romera2024mathematical} suggests that LLMs can outperform random operators by leveraging learned priors. A direct comparison with a non-LLM EA remains an important direction for future work.

\textbf{On morphological diversity.} A notable outcome of \method{} is the structural diversity of the discovered designs. Across terrains, the framework converges on qualitatively distinct morphologies, including wheeled quadrupeds on flat ground, high-clearance legged walkers on ridges, wide-stance hexapods on ice, and compact crawlers under beams. This diversity arises without any explicit diversity objective. We attribute this diversity to the LLM's broad prior over structural compositions, which allows it to propose fundamentally different body plans in a single generation, and to the terrain-specific fitness signal, which selectively reinforces the morphologies best suited to each environment. By contrast, grammar-based graph search typically converges on a narrower set of local optima because its mutation operators make only incremental structural changes. These diversity claims are qualitative rather than formally measured.

\textbf{Failure modes.} We observe three common failure modes: (1) grammar violations in early or zero-shot generations, which best-shot prompting largely mitigates, (2) kinematically degenerate designs (e.g., collapsed or overlapping joints) that compile but produce no useful motion, which are pruned by fitness-based selection, and (3) morphological convergence when prompt diversity is low, which our diverse initialization and fitness-score exclusion help counter.

%% file: sections/related_work.tex
\section{Related Work}
\label{sec:related_work}

\textbf{LLMs for evolutionary search.} LLMs have recently shown promise for creative engineering tasks~\cite{romera2024mathematical}, but their application to automated robot design remains limited~\cite{zeng2023large}. Early efforts introduced a human-in-the-loop framework that leverages ChatGPT to steer both conceptual and technical stages of design~\cite{stella2023can}, while \cite{makatura2023can} demonstrated how GPT-4 can synthesize manufacturable mechanical assemblies through discrete composition. Subsequent studies coupled LLMs with principled search or prompting strategies. \cite{guo2024connectinglargelanguagemodels} showed that combining evolutionary algorithms with LLMs produces powerful prompt optimizers, and \cite{wang2023grammar} introduced grammar prompting to ensure domain-specific language outputs remain syntactically valid. In parallel, LASeR~\cite{song2025laser} adopts a bi-level framework, using an LLM as a mutation operator to diversify robot morphologies within an evolutionary loop. Similarly, Text2Robot~\cite{ringel2024text2robot} translates natural-language descriptions into quadruped designs given a fixed limb count template. In contrast to these works, our method operates directly in a 3D design space and places no a priori limits on the number of limbs, enabling exploration of a far larger and more complex morphology search space.

\textbf{Robot design automation.} Beyond LLMs, robot design is shifting from exhaustive combinatorial search to data-driven methods that learn task-specific morphologies. Early work showed that deep reinforcement learning can act as a heuristic to guide the search for optimal modular designs~\cite{whitman2020modular}. Later, morphology generation was framed as a conditional generative problem using GANs to map task descriptions to diverse robot assemblies~\cite{hu2022modular}. GLSO~\cite{hu2023glso} extended this direction by embedding the discrete design space into a continuous latent space via a grammar-guided variational autoencoder. Beyond modular assemblies, diffusion models paired with learned dynamics networks have been used to generate custom end-effectors~\cite{xu2024dynamics}, and policy-gradient methods have been applied to voxel robots~\cite{li2024reinforcement}. In parallel, graph search has been augmented with neural heuristics~\cite{wang2019neural}, and~\cite{xu2021multi} proposed a multi-objective graph search that jointly optimizes morphology and control. Together, these advances have broadened the toolkit for automated robot design. Our work builds on this trend by incorporating an LLM as a generative model and an evolutionary pipeline to efficiently traverse the design space guided by performance feedback.

\textbf{Legged locomotion.} Classical control methods such as MPC~\cite{di2018dynamic, kim2019highly}, as used in Robogrammar, require accurate models and extensive tuning. We instead adopt model-free RL~\cite{hwangbo2019learning, haarnoja2018learning}, training a separate SAC~\cite{haarnoja2018soft} policy per design. Recent morphology-agnostic controllers~\cite{shafiee2024manyquadrupeds, bohlinger2025one} could replace per-morphology training in the future, and our modular framework is compatible with such advances.

%% file: sections/future_work.tex
\section{Limitations and Future Work} 
\label{sec:future_work}

\textbf{Expanding the design space.}
One future direction is to broaden the range of generated robot designs. This could be achieved by relaxing grammar constraints to permit a wider variety of materials, component types, and scalable link dimensions. Another promising avenue is to map grammar-defined components directly to hardware modules, enabling real-world assembly from a standardized kit of parts. 

\textbf{Diverse mix of environments.}
A natural next step is to extend the framework to robots that can traverse multiple terrain types, for example by initializing with few-shot examples tailored to different environments or incorporating textual terrain descriptions directly into the prompt. Achieving robustness across terrains would move closer to creating versatile, all-terrain robots and serve as a testbed for evaluating how well LLMs generalize design principles beyond single-task optimization.  

\textbf{Joint generation of a robot and its policy.} A promising direction is jointly generating robot morphologies and corresponding control strategies with LLMs. Recent studies, such as~\cite{ma2024eureka}, have shown that LLMs can generate reward functions without task-specific prompting or predefined templates. Building on this, LLMs could also propose high-level control heuristics or policy structures in parallel with morphology, which may help reduce the reliance on costly per-design RL training.

\textbf{Baselines and robustness.} Comparing against a traditional EA with random mutation and crossover on the same grammar and RL pipeline would isolate the LLM's contribution more precisely. Studying prompt sensitivity across variations in phrasing, few-shot formatting, and chain-of-thought instructions would clarify robustness to prompt design choices. Evaluating open-source LLM alternatives and incorporating quantitative diversity metrics would further strengthen the evaluation.

%% file: sections/conclusion.tex
\section{Conclusion} 
\label{sec:conclusion}
We presented \method{}, a framework that integrates LLMs, evolutionary algorithms, RL-based control, and robot grammars to automate the design of modular robots. Robot morphologies evolved by \method{} improve consistently over successive generations, and the final designs match or exceed the performance of Robogrammar's graph-search method across multiple terrains. The discovered morphologies, spanning wheeled quadrupeds, hexapods, and low-profile crawlers, demonstrate the LLM's capacity to explore beyond human-intuitive solutions while remaining physically feasible. Notably, these designs emerged without hand-crafted mutation operators or domain-specific heuristics, relying solely on the LLM's prior knowledge guided by evolutionary selection pressure.

More broadly, our results show that LLMs can serve as effective generative operators within evolutionary design loops, complementing rather than replacing traditional search. The key advantage lies in the LLM's ability to propose structurally complete and diverse candidates drawing on broad prior knowledge, reducing the need for manually designed variation operators. This paradigm extends naturally to other grammar-constrained design domains such as mechanisms, furniture, and architectural structures. In future work, we aim to extend \method{} beyond simulation by integrating real-world manufacturing constraints and rapid fabrication techniques to produce deployable, task-specific robots.

%% file: sections/acknowledgements.tex
\section*{Acknowledgments}
This work was partially supported by the National Science Centre, Poland, under PRELUDIUM Grant No.\ 2024/53/N/ST6/03370 and Sonata Bis Grant No.\ 2024/54/E/ST6/00388.